\documentclass[doublecolumn,oneside,letter]{IEEEconf}
\usepackage{amsmath}
\usepackage{amssymb}
\usepackage{graphicx}
\usepackage{subfigure}
\usepackage{verbatim}
\usepackage[thinlines,thiklines]{easybmat}
\usepackage{latexsym}
\usepackage[dvipsnames]{xcolor}
\usepackage{cite}
\usepackage{stmaryrd}
\usepackage{multirow}
\usepackage{textcomp}
\usepackage[ruled]{algorithm2e}
\usepackage{diagbox}
\usepackage{array}
\usepackage{bm}
\usepackage{color,soul}
\sethlcolor{yellow} %color set highlighting color \sethlcolor{white}

\def\s{\mathop{\rm s}\nolimits}
\def\c{\mathop{\rm c}\nolimits}
\newcommand{\bs}[1]{\ensuremath{{\boldsymbol{#1}}}}

\begin{document}

\title{\Large\bf Bipedal-Walking-Dynamics Model on Granular Terrains}

\author{Xunjie Chen\thanks{X. Chen, X. Huang, and J. Yi are with the Department of Mechanical and Aerospace Engineering, Rutgers University, Piscataway, NJ 08854 USA (email: xc337@rutgers.edu, xh301@scarletmail.rutgers.edu, jgyi@rutgers.edu).}, Xinyan Huang, Peter Shan\thanks{P. Shan is with Bridgewater Raritan High School, Bridgewater, NJ 08807 USA (email: {pshan2026@gmail.com}).}, Jingang Yi, and Tao Liu\thanks{T. Liu is with the State Key Lab of Fluid Power Transmission and Control and the School of Mechanical Engineering, Zhejiang University, Hangzhou 310027, China (e-mail: liutao@zju.edu.cn).}}

\maketitle
\thispagestyle{empty}
\pagestyle{empty}

\begin{abstract}
Bipeds have demonstrated high agility and mobility in unstructured environments such as sand. The yielding of such granular media brings significant sinkage and slip of the bipedal feet, leading to uncertainty and instability of walking locomotion. We present a new dynamics-modeling approach to capture and predict bipedal-walking locomotion on granular media. A dynamic foot-terrain interaction model is integrated to compute the ground reaction force (GRF). The proposed granular dynamic model has three additional degree-of-freedom (DoF) to estimate foot sinkage and slip that are critical to capturing robot-walking kinematics and kinetics such as cost of transport (CoT). Using the new model, we analyze bipedal kinetics, CoT, and foot-terrain rolling and intrusion affects. Experiments are conducted using a biped robotic walker on sand to validate the proposed dynamic model with robot-gait profiles, media-intrusion prediction, and GRF estimations. This new dynamics model can further serve as an enabling tool for locomotion control and optimization of bipedal robots to efficiently walk on granular terrains.
\end{abstract}

\vspace{-1mm}
\section{Introduction}
\label{sec:Intro}

Achieving rapid and efficient robotic locomotion on yielding terrains such as sand remains challenging for legged robots~\cite{AguilarRPP2016,GodonFRAI2023}. Unlike on rigid ground, the bipedal foot experiences significant sinkage and slippage when walking on granular terrains. Complex robot dynamics and foot-terrain interactions bring challenges in designing effective bipedal locomotion controllers. Although emerging techniques such as  data-driven, machine learning have shown promising results (e.g.,~\cite{choi2023learning}), obtaining high-fidelity training data from experiments or simulation are expensive~\cite{pragr2018cost}. Therefore, development of accurate robot dynamics on granular terrains is urgently needed for effective locomotion control and efficient-energy transport.

Terramechanic-based models, such as the Bekker model and Janosi formula, have been widely used to estimate the ground reaction forces (GRFs) and foot-terrain interactions on granular media~\cite{yangJTM2024comparative,YaoTMech2024}. An alternative class of resistive-force-theory (RFT) based models has also been developed~\cite{li2013terradynamics,aguilar2016robophysical,chen2025TMECH_Sand} for terrestrial legged robots in terms of stance-stability-region criterion design~\cite{XiongIROS2017}, optimal foot-shape design and gait and actuator efficiency~\cite{ChenICRA24}. However, few studies have examined how these models can be integrated with robot dynamics and incorporated with stable control synthesis for robotic locomotion.

For biped dynamics, simplified spring-mass models were used in~\cite{vejdani2013bio} to deal with robotic walking under unexpected ground-height disturbance and to design optimal locomotion synthesis. Centroidal dynamics and force-and-moment models~\cite{zhu2023BRAVER} were used for bipedal contact force deployment with the model-predictive-control framework. However, this class of reduced-order models only described the kinematics of the center of mass (CoM) and cannot be directly used for bipedal gait prediction. Other reduced-order models like actuated spring-loaded inverted pendulum (aSLIP) and hybrid-linear inverted pendulum (H-LIP) models~\cite{xiong2021slip} have been commonly used to provide an accurate dynamics approximation of gait.

Multi-link rigid body dynamics models have also been widely developed for analysis of humanoid gait considering roll dynamics of curved feet on rigid ground~\cite{martin2014predicting,rodman2021developing}. However, the contact holonomic constraints in previous work are still based on non-slip assumptions. To address the existence of slip explicitly, the previous assumption was relaxed and foot slip dynamics were built and predicted with the augmented 7-link model in~\cite{trkov2019bipedal}. The development in~\cite{trkov2019bipedal} further enabled the prediction of bipedal walkers under foot slip and recovery gaits~\cite{TrkovTASE2019}. Nevertheless, it only modeled planar dynamics on the rigid ground in the sagittal plane of the robot locomotion and no foot sinkage was considered.

On granular terrains, the non-slip and non-sinkage assumptions that were used in prior works are invalid due to the presence of liquid-like plastic flow~\cite{hamm2011dynamics} and solid-like effects such as local solidification and shear jamming~\cite{waitukaitis2012impact,chen2025TMECH_Sand}. In this work, we present an analytical dynamics model for the bipedal robot walking on granular terrain. The model introduces three additional degree-of-freedom (DoF) to deal with the estimation of vertical sinkage and horizontal slip experienced by the stance foot in media. Dynamics are decoupled in the sagittal and frontal planes of the bipedal robot. Integrated with a newly developed resistive force model~\cite{chen2025TMECH_Sand}, sinkage and slip dynamics are built. The proposed model is readily applied to other 6-link motor-actuated bipeds~\cite{zhu2023BRAVER,zhu2023proprioceptive,chen2025TMECH_Sand} and is validated by extensive walking experiments on sandy terrain. The proposed robot dynamics model potentially enables optimal gait design and simulation for machine-learning-based biped robot control on granular terrain.

\begin{figure*}[t!]
	\centering
	\subfigure[]{
		\includegraphics[width=1.9in]{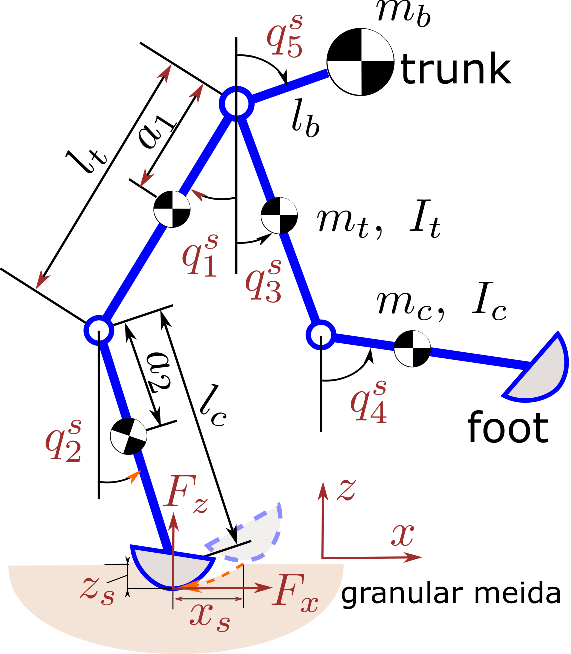}
		\label{fig:sagittal}}
\hspace{4mm}
	\subfigure[]
	{\includegraphics[width=1.88in]{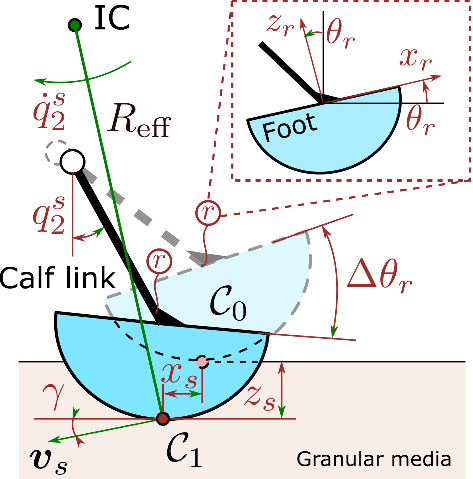}
		\label{fig:rollSchematics}}
\hspace{6mm}
	\subfigure[]
	{\includegraphics[width=1.85in]{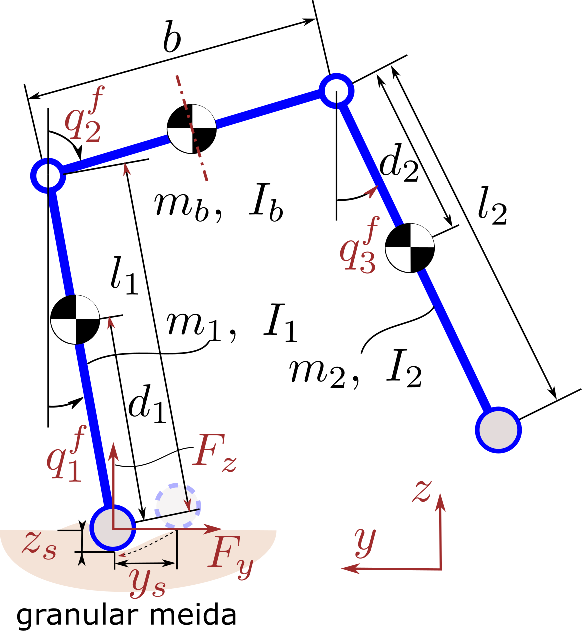}
		\label{fig:frontal}}
	\vspace{-1mm}
	\caption{Schematic of the bipedal walking model with foot sinkage and slip on granular media. (a) Sagittal  dynamics. (b) Stance foot rolling and sinkage with a convex foot shape in the sagittal plane. (c) Frontal dynamics.}
	\vspace{-3mm}
\end{figure*}

The main contributions of this work are twofold. First, for the first time, this work extends the bipedal dynamics model with foot sinkage and slip on yielding terrains. Unlike existing work that modeled soft terrain with elasticity~\cite{fahmi2020stance}, an accurate foot-terrain interaction model is integrated to capture yielding terrain properties. As a result, the proposed model reliably estimates GRFs, transient rolling and slip effects, and cost of transport (CoT) over granular media. This insightful development enables advanced balancing locomotion control and simulation on granular terrains. Second, compared with the aSLIP models in~\cite{xiong2021slip,xiong2022TRO} and planar humanoid dynamics in~\cite{rodman2021developing,trkov2019bipedal}, the proposed model offers multiple advantages such as accurate predictions of joint kinematics and whole-body work energetics with low computational cost.

%The remainder of the paper is organized as follows: We first present the bipedal robot dynamics in Section~\ref{sec:Model}. The experimental setup and description are discussed in Section~\ref{sec:Exp}. Experimental results and discussion are presented in Section~\ref{sec:result}. Finally, we summarize the conclusions in Section~\ref{sec:Cons}.

\section{Biped Dynamics Model on Granular Terrains}\label{sec:Model}
\subsection{Sagittal Dynamics}
The biped walker is modeled as a five-link system. Fig.~\ref{fig:sagittal} illustrates the modeling schematic with the single stance leg submerging in the granular terrain. The trunk has the concentrated mass and each leg consists of the thigh and calf links. The bilateral leg configuration of the sagittal dynamics is symmetrical. We select absolute angles with respect to the vertical direction of two legs and trunk as coordinates $\bs{q}^s = [q^s_1\;q^s_2\;q^s_3\;q^s_4\;q^s_5]^{\top}\in \mathbb{R}^5$, where the superscript ``s'' represents variables in the sagittal plane. Only $q_i^s$, $i=1,2,3,4$, are actuated by joint torques $\bs{\tau}_s = [\tau_1^s \;\tau_2^s \;\tau_3^s \;\tau_4^s]^{\top}\in\mathbb{R}^4$.

To capture the dynamics in the granular media, we introduce three additional intrusion variables. Fig.~\ref{fig:rollSchematics} shows the schematic of the stance foot-terrain interactions. Convex foot shape is considered for rolling-slip analysis (fixed at the end of the calf link) and the contact point is assumed as the lowest point in the media. We denote the vertical intrusion, longitudinal slip and lateral slip distances of stance foot contact point $\mathcal{C}_1$ by $z_s$, $x_s$, and $y_s$, respectively. These values are calculated from initial contact location $\mathcal{C}_0$. With these variables, the augmented coordinates are defined as $\bs{q}_s = [(\bs{q}^s)^{\top} \; x_s\; z_s]^{\top}\in \mathbb{R}^7$.

The single-stance dynamics in the sagittal plane are described as
\begin{equation}
\label{eqn:sagittal}
\bs{D}_s(\bs{q}_s)\ddot{\bs{q}}_s + \bs{C}_s(\bs{q}_s,\dot{\bs{q}}_s)\dot{\bs{q}}_s + \bs{G}_s(\bs{q}_s) = \bs{B}_s\bs{\tau}_s + \bs{J}^{\top}_{s}\bs{F}_s,
\end{equation}
where $\bs{D}_s(\bs{q}_s)$, $\bs{C}_s(\bs{q}_s,\dot{\bs{q}}_s)$, and $\bs{G}_s(\bs{q}_s)$ the inertia, Coriolis, gravity matrices, respectively, for the sagittal plane. The actuation torque mapping matrix $\bs{B}_s=[\bs{I}_4,\bs{0}]^{\top} \in \mathbb{R}^{7\times 4}$, where $\bs{I}_N$, $N \in \mathbb{N}$, is an $N$-dimensional identity matrix. For the stance foot in granular media, the reaction force resulting from the foot-terrain interactions is denoted by $\bs{F}_s = [F_x \, F_z]^{\top}$ and the associated Jacobian matrix is $\bs{J}_{s} = [\partial x_s(\bs{q}_s)/\partial \bs{q}_s, \partial z_s(\bs{q}_s)/\partial \bs{q}_s]^{\top}$.

Explicitly, we expand the last two rows of the sagittal dynamics \eqref{eqn:sagittal} that are related to two extra DoFs, namely, $x_s$ and $z_s$. For $x_s$ dynamics, we have
\begin{equation}
\label{eqn:ddxs}
  M_s\ddot{x}_s +  g_5(\dot{q}_5^s,\ddot{q}_5^s) + g_1(\dot{q}_1^s,\ddot{q}_1^s) + g_2(\dot{q}_2^s,\ddot{q}_2^s) = F_x,
\end{equation}
where $M_s = m_b + m_t + m_c$, the total mass of the trunk, thigh, and calf links. The other three functions of absolute angles $q_i^s$, $i = 1,2,5$, are $g_1(\dot{q}_1^s,\ddot{q}_1^s)= (m_ta_1 + m_cl_t)[-\c_{q_1^s}\ddot{q}_1^s+\s_{q_1^s}(\dot{q}_1^s)^2]$, $g_2(\dot{q}_2^s,\ddot{q}_2^s)= m_ca_2[-\c_{q_2^s}\ddot{q}_2^s+\s_{q_2^s}(\dot{q}_2^s)^2]$, and $g_5(\dot{q}_5^s,\ddot{q}_5^s)= m_b l_b \c_{q_5^s} \ddot{q}_5^s - m_b l_b \s_{q_5^s}(\dot{q}^s_5)^2$, where notations $\c_{q_i^s} = \cos{q_i^s}$ and $\s_{q_i^s} = \sin{q_i^s}$ are used for $q_i^s$, $i = 1,2,5$ and other joint angles in this paper.

Similarly, for the vertical intrusion $z_s$, we express
\begin{equation}\label{eqn:ddzs}
  M_s\ddot{z}_s +  h_5(\dot{q}_5^s,\ddot{q}_5^s) + h_1(\dot{q}_1^s,\ddot{q}_1^s) + h_2(\dot{q}_2^s,\ddot{q}_2^s) = F_z-M_sg,
\end{equation}
where $h_{1,2,5}(\dot{q}_i^s,\ddot{q}_i^s)$ are listed as follows: $h_1(\dot{q}_1^s,\ddot{q}_1^s)= (m_ta_1 + m_cl_t) [\s_{q_1^s} \ddot{q}_1^s+\c_{q_1^s}(\dot{q}_1^s)^2]$, $h_2(\dot{q}_2^s,\ddot{q}_2^s)= m_ca_2[\s_{q_2^s}\ddot{q}_2^s+\c_{q_2^s}(\dot{q}_2^s)^2]$, and $h_5(\dot{q}_5^s,\ddot{q}_5^s)=-m_b l_b \s_{q_5^s} \ddot{q}_5^s - m_b l_b \c_{q_5^s}(\dot{q}^s_5)^2$. With given reaction force measurements or foot-terrain interaction model, from \eqref{eqn:ddxs} and \eqref{eqn:ddzs}, the dynamic accelerations are updated in the granular media and estimate corresponding intrusion variables.

The intrusive velocity of the foot in the sagittal plane is denoted as $\bs{v}_s=[\dot{x}_s \; \dot{z}_s]^\top$ at the contact point and the corresponding velocity directional angle is denoted by $\gamma$, that is, $\gamma=\tan^{-1}\left(\dot{z}_s/\dot{x}_s\right)$. For rigid-ground walking, $\dot{z}_s\equiv0$ so that $\gamma\equiv 0$. Fig.~\ref{fig:rollSchematics} illustrates a local rolling frame $x_r o_r z_r$. The corresponding orientation angle of the stance foot with respect to the horizontal direction is denoted by $\theta_r$ and then the rolling angle of the foot from the beginning of the stance is denoted by $\Delta \theta_r$.
$\theta_r$ is calculated through the actual contact point in the frame $x_r o_r z_r$. Given the foot shape contour $z_r = \mathcal{S}(x_r)$ in the local frame and $\mathcal{C}_1$ as the lowest point of the foot, the $x_r$-component of $\mathcal{C}_1$ satisfies:
\begin{equation*}
  \frac{x_r}{\mathcal{S}(x_r)} = \frac{d\mathcal{S}(x_r)}{dx_r}.
\end{equation*}
Then, the instant rolling angle $\Delta \theta_r$ between $\mathcal{C}_1$ (with $\theta_{r1}$) and $\mathcal{C}_0$ (with $\theta_{r0}$) is given by
\begin{equation}
\label{eqn:theta_r}
  \Delta \theta_r = \theta_{r1}-\theta_{r0}, ~\tan{\theta_{ri}} = \frac{dz_r}{dx_r} \bigg |_{\mathcal{C}_i},~i=0,1,
\end{equation}
where $\theta_{r0}$ and $\theta_{r1}$ are the foot orientation angles at the initial contact and current contact instance, respectively.
The instantaneous center (IC) is associated with $\mathcal{C}_1$ and corresponding effective rolling radius $R_{\mathrm{eff}}$ is defined as
\begin{equation}
\label{eqn:Reff}
R_{\mathrm{eff}} \approx \frac{\|\bs{v}_s\|}{|\dot{\theta}_r|}=\frac{\dot{x}_s \sqrt{1+\tan^2{\gamma}}}{|\dot{\theta}_r|}.
\end{equation}

\subsection{Frontal Dynamics}

Fig.~\ref{fig:frontal} illustrates the schematic of the bipedal walker motion in the frontal plane with the lateral foot slip. The biped is modeled as a three-link rigid body. We also select the absolute angles of each link with respect to the vertical direction as generalized coordinates $\bs{q}^f = [q^f_1\,q^f_2\,q^f_3]^{\top}\in \mathbb{R}^3$, where the superscript ``f'' represents in the frontal plane. There are two actuation torque inputs (for $q_2^f$ and $q_3^f$) at the two hip joints and we denote the joint torque as $\bs{\tau}_f = [\tau_2^f \,\tau_3^f]^{\top}\in \mathbb{R}^2$. Similarly, we choose the augmented coordinates by including $y_s$ and $z_s$ such that $\bs{q}_f = [(\bs{q}^f)^{\top} \; y_s \; z_s]^{\top}\in \mathbb{R}^5$. Therefore, the single-stance dynamics in the frontal plane of the robot are expressed by
\begin{equation}
\label{eqn:frontalDyn}
\bs{D}_f(\bs{q}_f)\ddot{\bs{q}}_f + \bs{C}_f(\bs{q}_f,\dot{\bs{q}}_f)\dot{\bs{q}}_f + \bs{G}_f(\bs{q}_f) = \bs{B}_f\bs{\tau}_f + \bs{J}^{\top}_{f}\bs{F}_f,
\end{equation}
where $\bs{D}_f(\bs{q}_f)$, $\bs{C}_f(\bs{q}_f,\dot{\bs{q}}_f)$, and $\bs{G}_f(\bs{q}_f)$ are the inertia, Coriolis, gravity matrices in the frontal plane, respectively. The actuation-torque mapping matrix $\bs{B}_f=[\bs{0} \;\bs{I}_2 \;\bs{0}]^{\top} \in \mathbb{R}^{5\times 2}$. The external reaction force is $\bs{F}_f = [F_y\; F_z]^{\top}$ and the corresponding Jacobian matrix is defined as $\bs{J}_{f} = [\partial y_s(\bs{q}_f)/\partial \bs{q}_f \; \partial z_s(\bs{q}_f)/\partial \bs{q}_f]^{\top}$. The lateral slip dynamics are given by
\begin{equation}
\label{eqn:ddys}
    M_f\ddot{y}_s +  f_1(\dot{q}_1^{f},\ddot{q}_1^f) + f_2(\dot{q}_2^f,\ddot{q}_2^f) + f_3(\dot{q}_3^{f},\ddot{q}_3^f) = F_y,
\end{equation}
where $M_f = m_b + m_1 + m_2$, the total mass of the trunk and two legs. We also express the three functions of the absolute angles $q_i^f$, $i = 1,2,3$, by $f_1(\dot{q}_1^f,\ddot{q}_1^f)=-(m_1d_1 + m_bl_1 + m_2l_1)[\c_{q_1^f} \ddot{q}_1^f -\s_{q_1^f}(\dot{q}^f_1)^2]$, $f_2(\dot{q}_2^f,\ddot{q}_2^f)= (\frac{1}{2}m_bb + m_2b) [\c_{q_2^f}\ddot{q}_2^f-\s_{q_2^f}(\dot{q}_2^f)^2]$, and $f_3(\dot{q}_3^f,\ddot{q}_3^f)=m_2d_2[\c_{q_3^f}\ddot{q}_3^f-\s_{q_3^f}(\dot{q}_3^f)^2]$.

\subsection{Foot-Terrain Interactions}
The 3-DoF foot-terrain interaction model in~\cite{chen2025TMECH_Sand} is used for a computationally efficient approach by providing closed forms for the resistance forces in the presence of the intrusion variables. In the sagittal plane of the stance foot, the foot-terrain interaction model considers the two-dimensional (vertical and longitudinal) propagation dynamics of impact-induced solidification zone, which is treated as an added mass of the robot foot. Therefore, the vertical and longitudinal resistive forces $F_x$ and $F_z$ are computed. In the frontal plane, the bulldozing resistance is modeled for the main source of the lateral force $F_y$. {\color{black}A semi-cylindrical foot is taken for rolling contact.} Note that the closed forms of reaction forces are relative to the local solidification zone and the robot foot shapes result in the corresponding variations~\cite{ChenICRA24,chen2025TMECH_Sand}.

The closed forms of the resistive forces in the sagittal plane are given by~\cite{chen2025TMECH_Sand}
\begin{equation}
\label{eqn:FxFz}
  F_j = \frac{\alpha_i(\bs{n}(\phi_s),\bs{v}_s)}{2\tan{\phi_s}}WI_{z_s}, \;\; i = x,z,
\end{equation}
where $W$ is the foot width and $\bs{n}(\phi_s)$ is the solidification zone contour normal vector related to $\phi_s$. The granular fraction angle $\phi_s \approx 38^{\circ}$ is assumed in general for most sands~\cite{chen2025TMECH_Sand}. In~\eqref{eqn:FxFz}, $\alpha_i(\bs{n},\bs{v}_s)$, $i = x,z$, are the local stresses per unit length in the longitudinal and vertical directions, respectively~\cite{li2013terradynamics}. The term $I_{z_s}$ is a function of intrusion depth~$z_s$. The lateral bulldozing force is calculated as
\begin{equation}\label{eqn:Fy}
  F_y = \lambda\left(1-e^{-y_s/\lambda}\right)\alpha_y\left(\bs{n}_{\shortparallel},\bs{v}_f\right)g_{z_s},
\end{equation}
where $\alpha_y(\bs{n}_{\shortparallel},\bs{v}_f)$ is a local stress per unit length in the lateral direction, and $\bs{v}_f=[\dot{y}_f \; \dot{z}_f]^\top$ is the intrusive velocity in the frontal plane. $\bs{n}_{\shortparallel}$ indicates the normal vector of the foot side and that is parallel to the horizontal direction. In \eqref{eqn:Fy}, the coefficient $\lambda$ is the saturation length and $g_{z_s}$ is a function of
intrusion depth $z_s$. Details of the derivations in terms of ground reaction forces are found in~\cite{chen2025TMECH_Sand}.
\begin{figure*}[t!]
	\centering
	\includegraphics[width=6.7in]{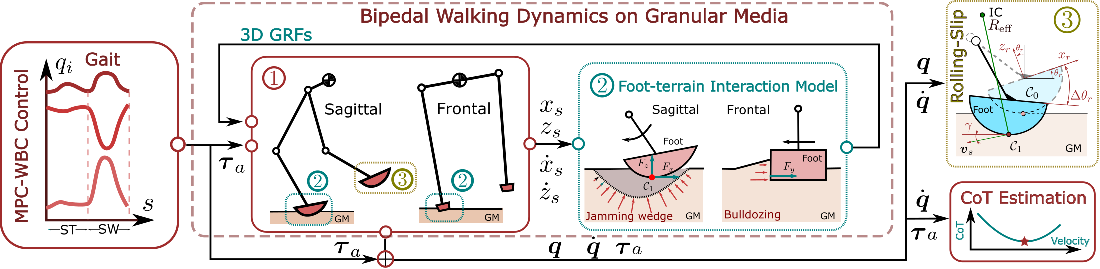}
	\caption{{Bipedal-walking-dynamics modeling on granular media with the inclusion of foot–terrain interactions, the rolling-slip effect, and the energy efficiency in terms of CoT.}}
	\label{fig:flowChart}
	%\vspace{-1mm}
\end{figure*}

Fig.~\ref{fig:flowChart} illustrates the inter-connection relationship between the augmented robot dynamics and the foot-terrain interaction model. With the foot-intrusion motion estimation ($x_s$, $y_s$, and $z_s$ and  their velocities), robot dynamics and terrain mechanics are bidirectionally coupled; the gait motion influences foot penetration, while granular resistive forces feed back into overall robot dynamics. By integrating foot-shape configuration, the robot dynamics are used for the rolling analysis, providing a geometric and kinematic basis to quantify how granular media deforms. This allows to bridge the gap of robot locomotion on granular media and rigid ground. Furthermore, the inclusion of foot–terrain interactions and rolling-slip effects ensures that the energy-expenditure estimation from the body dynamics model reflects implications of the additional energetic penalties imposed by granular substrates, such as increased resistive losses and reduced effective step efficiency.

\begin{figure*}[t!]
	\centering
	\subfigure[]{\includegraphics[width=3.7in]{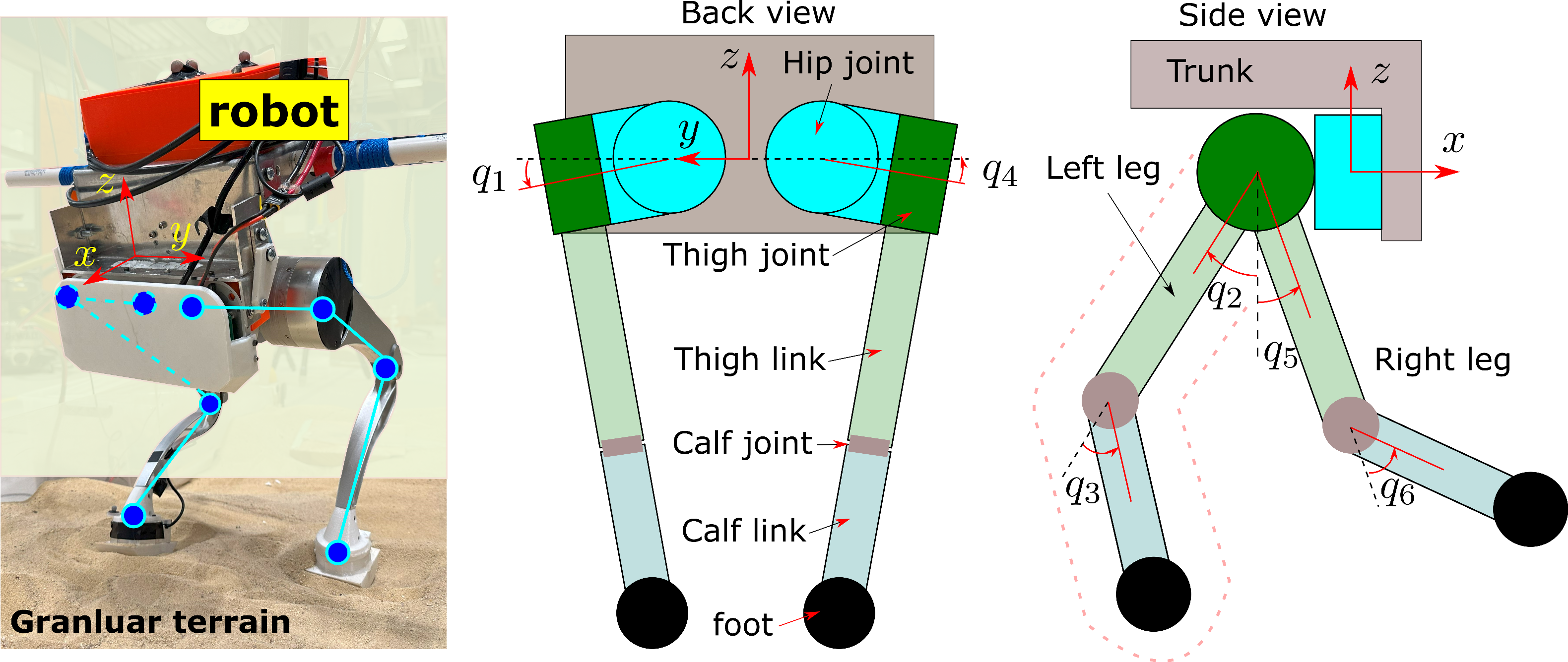}
		\label{fig:biped}}
	\subfigure[] {\includegraphics[width=3.1in]{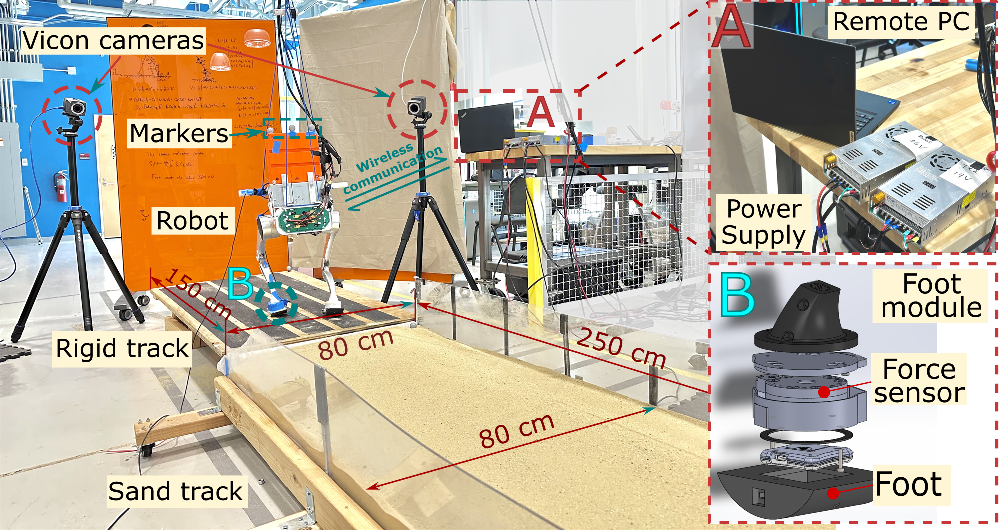}
		\label{fig:expSetup}}
	\caption{Experimental setup of the bipedal robot walking locomotion on granular media. (a) Schematic of the actuated joint configuration of the bipedal robot on granular media. (b) A bipedal robot on the sand track with a modularized foot.}
	\label{fig:Exp}
\end{figure*}

\section{Experiments}\label{sec:Exp}
\subsection{Bipedal Robot and CoT Calculation}
Fig.~\ref{fig:biped} shows the bipedal robot used in experiments. The robot (model YoboGo-10S from Yobotics Inc., China) is considered as a 6-link rigid body with two hip joints/links, two thigh joints/links, two calf joints/links, and a trunk. The robot has six actuation motors on each joint. We denote the relative angles of the six joint actuators by $\bs{q}_a = [q_1 \, \cdots\,q_6]^{\top}$ and the associated actuation torques by $\bs{\tau}_a = [\tau_1 \, \cdots \, \tau_6]^{\top}$. The bipedal robot was controlled through an onboard computer (Jetson TX2 from Nvidia Corp.). The actuation joint angle ($\bs{q}_a$), angular velocity ($\dot{\bs{q}}_a$) and acceleration ($\ddot{\bs{q}}_a$), and torque feedback ($\bs{\tau_a}$) were collected by the encoders and motor current measurements for the locomotion gait.

For the configuration, we assign the joint angles $q_i$, $i = 1,2,3$, for the left leg and $q_j$, $j = 4,5,6$, for the right leg, respectively. The walking gait of the bipedal robot consists of repeated single-stance and swing phases for each leg. We denote the leg-joint angle in the single-stance phase and swing phase as $\bs{q}_{st}$ and $\bs{q}_{sw}$, respectively. We do not consider the double-stance phase for walking locomotion.
The conversions between the robot actuation coordinates $\bs{q}_a$ and the model joint coordinates $\bs{q}^s$ and $\bs{q}^f$ are provided for both the sagittal and frontal planes. For the sagittal plane, we have $\bs{q}_a = \bs{S}_s\bs{q}^s$ where $\bs{S}_s\in \mathbb{R}^{6\times5}$ is a constant mapping matrix. The associated actuation torque input for the sagittal dynamics in~\eqref{eqn:sagittal} is computed as $\bs{\tau}_s = \frac{\partial \bs{q}_a}{\partial\bs{q}^s}\bs{\tau}_a$ such that $\tau_i^s = \sum_{j = 1}^{6}\frac{\partial q_j}{\partial q_i^s}\tau_j, ~i = 1,2,3,4.$
For the frontal dynamics in~\eqref{eqn:frontalDyn}, the relative angles of two robot hips $q_1$ and $q_4$ are expressed as $q_1 = -q_1^f + q_2^f$ and $q_4 = \pi - q_2^f + q_3^f$, respectively. Therefore, the actuation torque inputs of the frontal dynamics are $\tau_2^f = \tau_1-\tau_4$ and $\tau_3^f = \tau_4$.

%Note that the equivalent leg length in the frontal plane is revised by
%\begin{equation*}
%  l_i = \sqrt{l_t^2 + l_c^2 - 2l_t l_c \cos{(\pi-q_{3i})}}, \;\;i = 1,2,
%\end{equation*}
%where $l_t$ and $l_c$ are the length of the robot thigh link and calf link, respectively.

The CoT is a fundamental performance metric in terms of energetics for bipedal robot locomotion. The CoT is significant since forces and energy are easily dissipated in granular media. In this work, CoT is defined as the energy required to support a unit weight of the bipedal robot for moving a unit distance and therefore. We write CoT as $\mathrm{CoT} = \frac{E}{W_rd} =	\frac{1}{W_rd}\int_{t_0}^{t_f}\|\bs{\tau}_a^{\top}(t)\bs{\dot{q}}_a(t)\|dt$, where $E$ is total energy consumed during the locomotion, namely, mainly mechanical work done by the joint actuators, and $W_r$ is the total weight of the robot. For the duration, $t_0$ and $t_f$ are the start and terminal timings of the locomotion task, respectively, and $d$ is the robot walking distance which is obtained by integrating the longitudinal velocity of the robot CoM velocity $v_{\text{CoM}}^x(t)$. By decoupling the sagittal and frontal dynamics, we also provide the CoT estimation by $\mathrm{CoT} = \frac{1}{W_rd}\sum_{i=s,f}\int_{t_0}^{t_f}\|\left(\bs{B}_i\bs{\tau}_i(t)\right)^{\top}\bs{\dot{q}}^i(t)\|dt$.

\begin{figure*}[t!]
	\centering
	\subfigure[] {\includegraphics[width=2.50in]{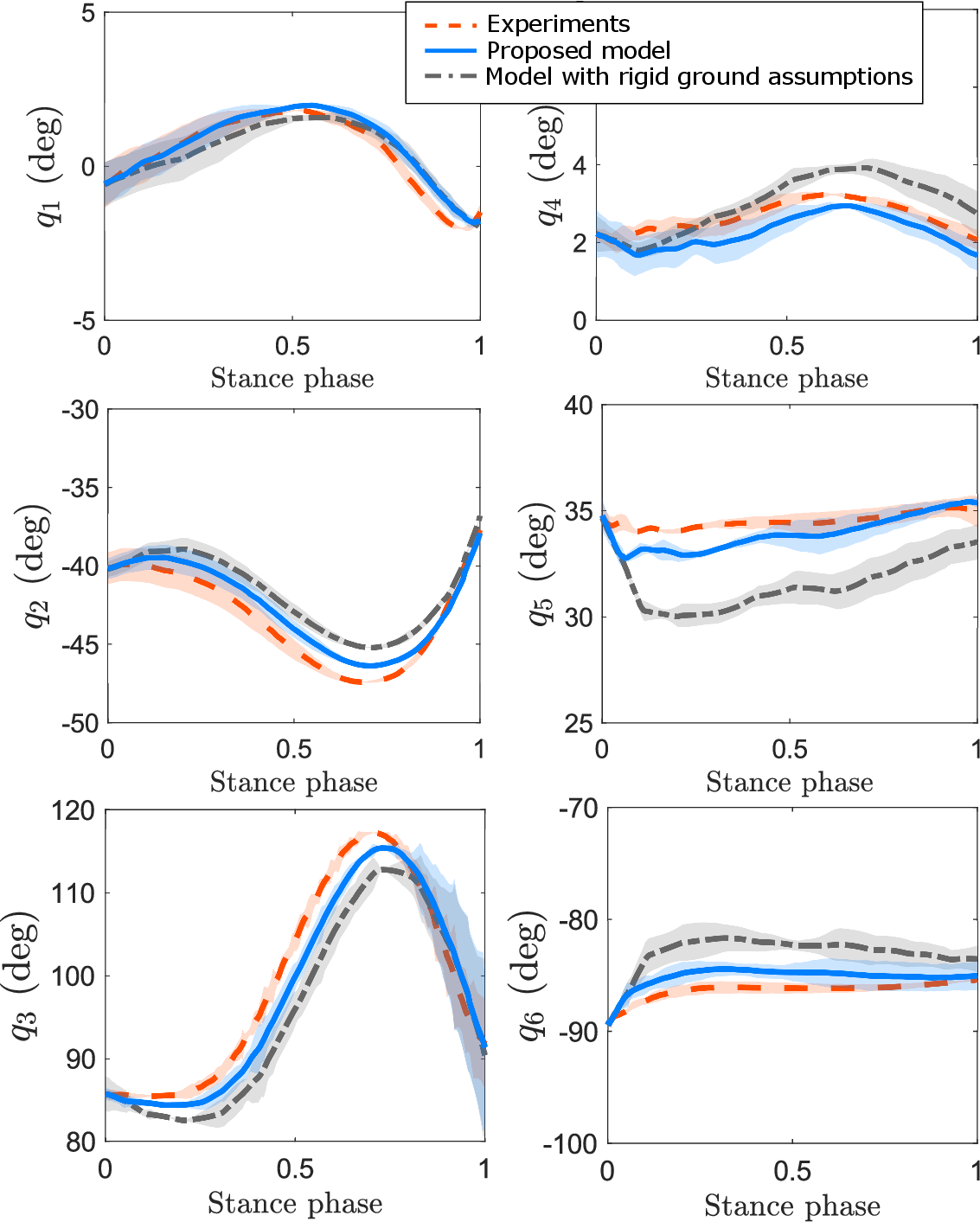}
		\label{fig:q}}
\hspace{0mm}
	\subfigure[] {\includegraphics[width=2.0in]{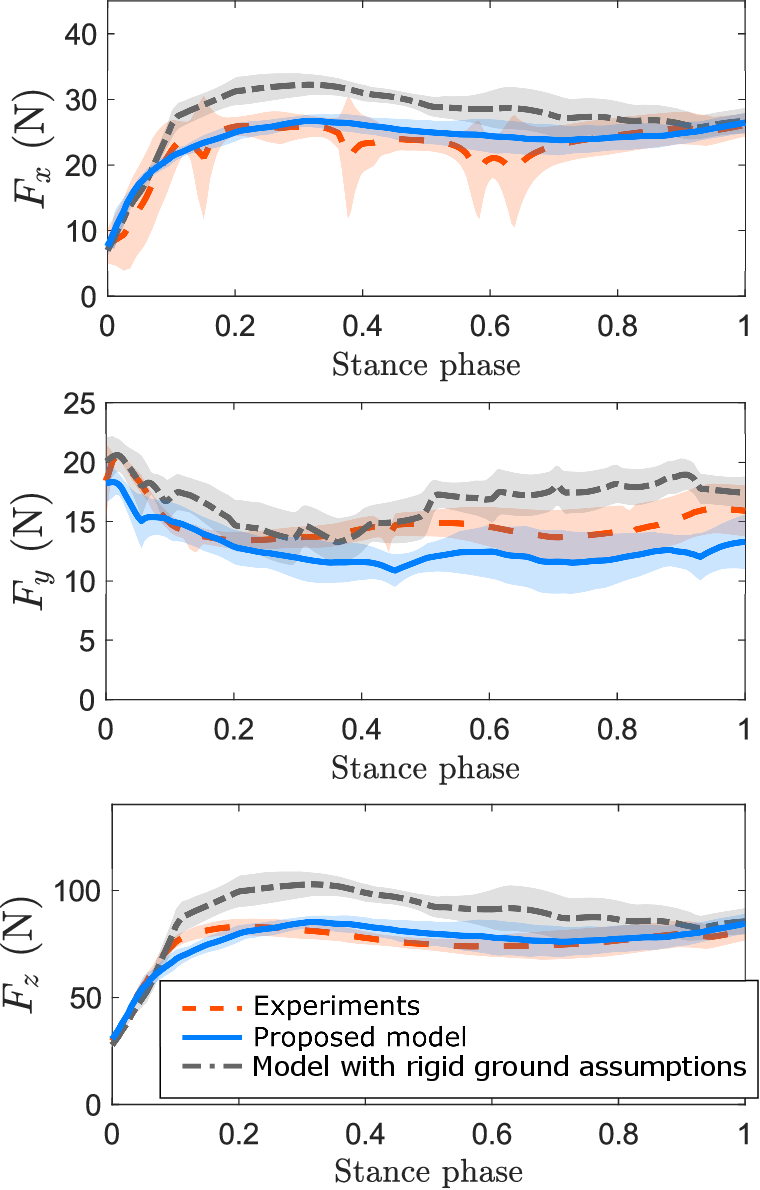}
		\label{fig:FxFyFz}}
\hspace{0mm}
	\subfigure[] {\includegraphics[width=2.01in]{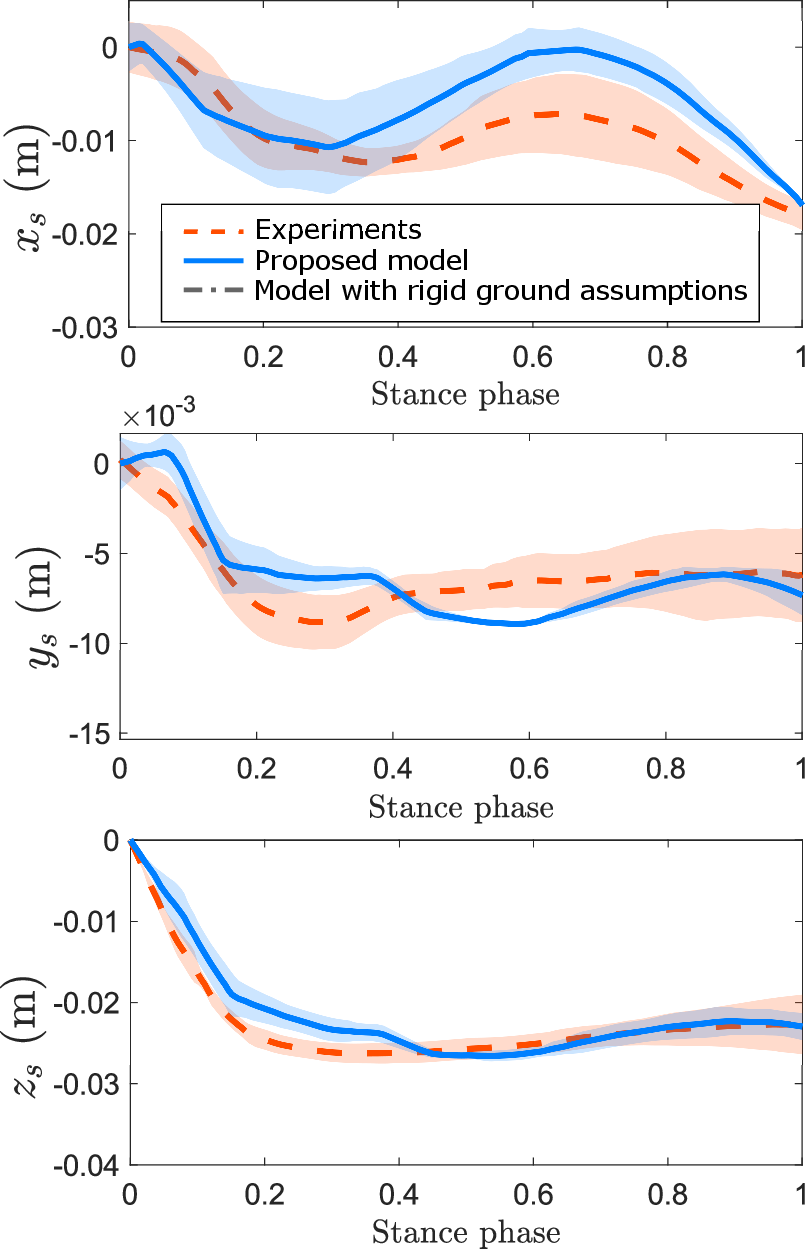}
		\label{fig:xys_sand}}
	\caption{Comparisons of robot locomotion results on granular media within one stance phase. Shadowed areas represent one standard deviation. (a) The gait comparison of the model prediction and experiment results. Swing leg: $q_{1,2,3}$, stance leg: $q_{4,5,6}$. (b) GRFs on the granular media. (c) Stance foot intrusion variables $x_s$, $y_s$ and $z_s$.}
	\label{fig:Locomotion}
\end{figure*}

\subsection{Experimental Setup}
Fig.~\ref{fig:expSetup} shows the laboratory setup of the bipedal robot walking locomotion on granular media. For walking locomotion, optical markers were attached to the robot body to capture corresponding kinematic information through a motion-capture system (8 Bonita cameras from Vicon Inc.). An indoor walking track ($L\times W=400\times 80$~cm) was designed for continuous walking~\cite{zhu2025JBE}.
%The track consisted of two segments, one for rigid ground and the other for sandy terrain. The thickness of the sand layer in this work was 12~cm throughout the entire sand segment, the same height of the rigid segment.
A customized semi-cylindrical shape was used for the modularized robot foot and one force sensor (Mini45 from ATI Inc.) was embedded for measuring GRFs.
%The embedded force sensor in the robot foot was first calibrated in association with force plate measurements to make sure that the accurate external force acting on the foot was captured. The robot was controlled to step on the force plate continuously and it maintained the body position without significant variations. The local force measurements from the force sensor in the foot frame were transferred to the world frame where the force plate was located~\cite{zhu2025JBE}. The robot body orientation matrix with respect to the world frame was obtained from the motion capture system and the orientation forward kinematic mapping from the body to the foot frame was calculated from the encoders.
For the granular media, we selected natural sand with average particle diameter 0.6-1.1~mm and mass density 1.66~g/cm$^3$. Vertical and horizontal penetration tests using a small plate were conducted to calibrate the terramechanic parameters of granular media (i.e., a scaling factor $\zeta=1.36$ for the local stress terms $\alpha_{i}$ ($i = x, y,z$)~\cite{li2013terradynamics,chen2025TMECH_Sand}) and the bulldozing resistive parameter $\lambda$, respectively. Other model parameters of the sagittal and frontal dynamics were computed by the accurate biped URDF file.

In walking experiments, the biped was first required to walking in place on the rigid platform until the gait became stable and consistent. Then, the biped was commanded to step forward crossing the rigid segment and sand track with a constant moving velocity {\color{black}using the model predictive control and whole-body control (MPC-WBC) framework~\cite{MihalecTMech2023}}. Before each walking trial, the granular media was manually blended to be uniform throughout the entire sand track and the sand surface was also leveled. The sand level height (i.e., $12$~cm) was known in advance. During the walking, the position of the robot was collected and relative foot position was calculated by forward kinematics of the biped.
%Therefore, the experimental foot intrusion velocity was calculated by $\bs{v}_{\text{int}}=[\dot{x}_s \,\dot{y}_s \,\dot{z}_s]^{\top} = \bs{v}_{\text{CoM}} + {}^{W}\!\mathbf{R}^{B}[\bs{\omega}^{B}]_{\times}\bs{h}_{\text{CoM}}(\bs{q}_{st})-{}^{W}\!\mathbf{R}^{B}\bs{J}_{\text{CoM}}(\bs{q}_{st})\dot{\bs{q}}_{st}$, where $[\bs{\omega}^B]_{\times}$ is the skew-symmetric matrix of the angular velocity of the robot body, $\bs{h}_{\text{CoM}}(\bs{q}_{st})$ and $\bs{J}_{\text{CoM}}(\bs{q}_{st})$ are the forward kinematic mapping and Jacobian matrix of the stance leg with respect to the CoM, respectively.
We obtained the intrusion variables by integration for comparisons.

\setcounter{figure}{5}
\begin{figure*}[t!]
	\centering
	\subfigure[] {\includegraphics[width=1.66in]{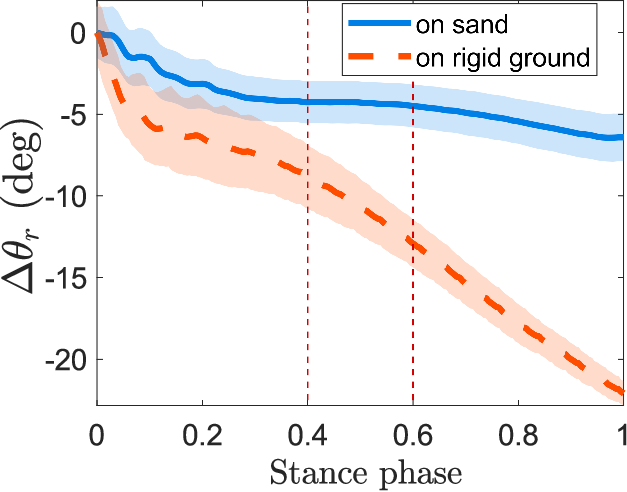}
		\label{fig:rolling_theta}}
	\subfigure[] {\includegraphics[width=1.62in]{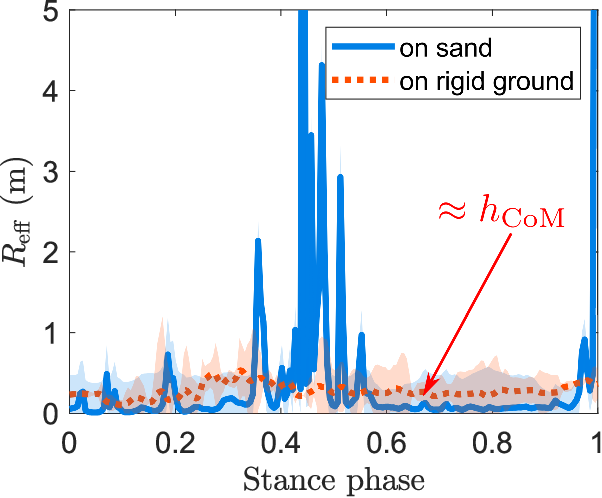}
		\label{fig:rolling_Reff}}
	\subfigure[] {\includegraphics[width=1.62in]{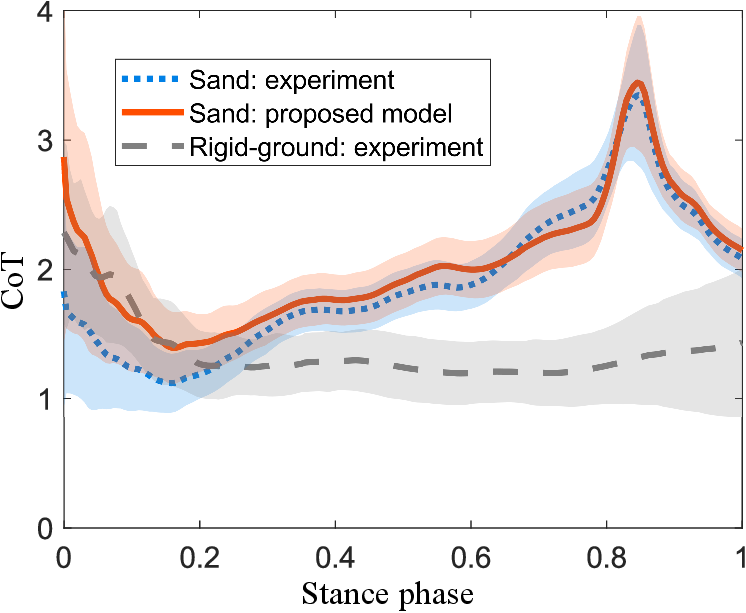}
		\label{fig:CoT_stance}}
	\subfigure[] {\includegraphics[width=1.67in]{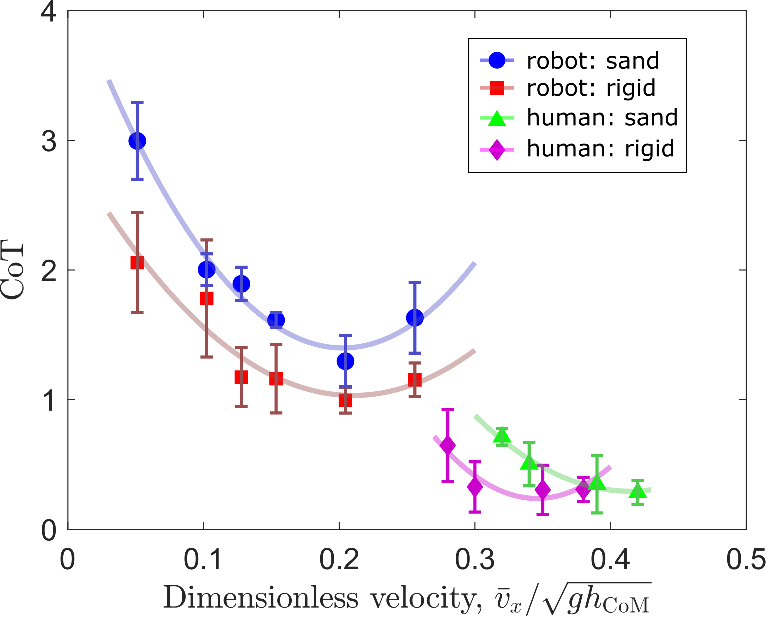}
		\label{fig:CoT_velocity}}
	\caption{(a)-(c) Foot orientation angle $\Delta \theta_r$ (negative sign presents decreasing values), the effective rolling radius of the stance foot, and the CoT of walking on sand and rigid ground during one stance phase, respectively. Shadowed areas represent the standard deviations. $\bar{v}_{\mathrm{CoM}}^x/\sqrt{gh_{\mathrm{CoM}}}=0.1$. (d) Velocity dependency of locomotion CoT of the bipedal robot and human on sand and rigid ground, respectively. Each CoT value is mean$\pm$standard-variation. The solid lines are second-order polynomial fitting. CoT of humanoid locomotion are based on the data by\cite{zhu2025JBE}.}
	\label{fig:ExpResults}
\end{figure*}

\setcounter{figure}{4}
\begin{figure}[h!]
  \centering
  	\subfigure[] {\includegraphics[width=3.32in]{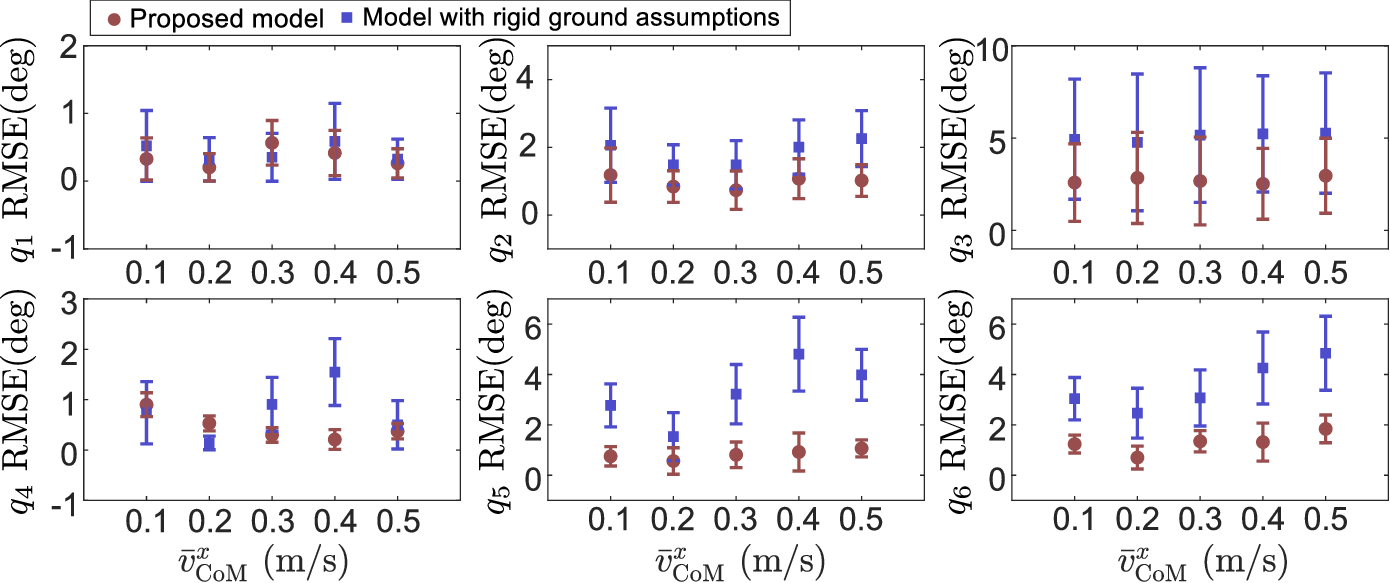}
		\label{fig:RMSE_q}}
  	\subfigure[] {\includegraphics[width=3.32in]{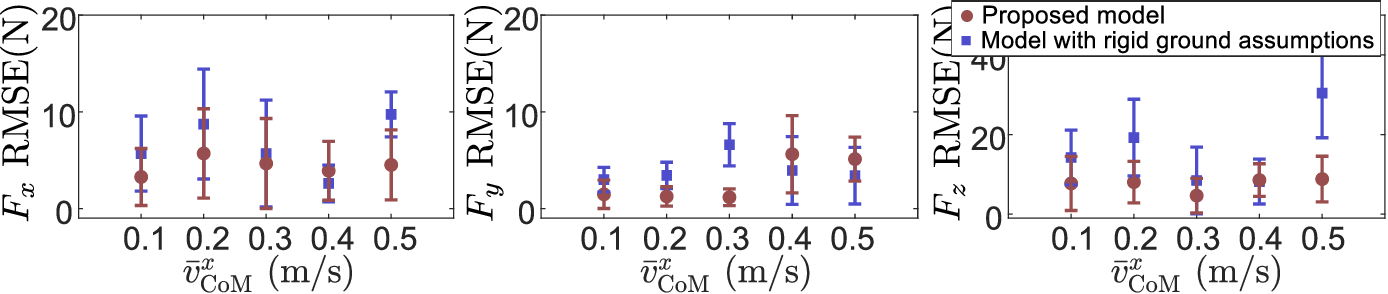}
		\label{fig:RMSE_GRFs}}
  	\subfigure[] {\includegraphics[width=3.32in]{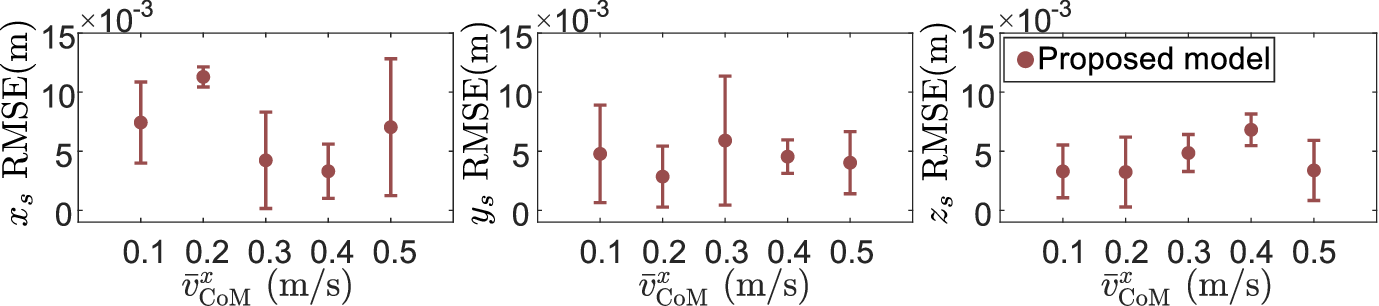}
		\label{fig:RMSE_xyz}}
  \caption{{\color{black}Comparisons of RMSE under different forward-velocity commands. (a) Joint angle $\bs{q}$. (b) GRFs. (c) Intrusion variables $x_s$, $y_s$, and $z_s$. Circle dots: the proposed model. Square dots: model with rigid-ground assumptions.}}\label{fig:rmseVel}
  \vspace{-3mm}
\end{figure}

The walking gait was consistent and swing/stance timing was known in advance.
The total gait cycle period was $0.4$~s with $50\%$ swing phase and $50\%$ stance phase. The trajectory of the swing foot was designed as a cycloid in terms of the swing phase given the foot-placement proportional to the forward moving velocity and the fixed swing height (i.e., $10$~cm).
We conducted experiments with various robot moving velocities. In the experiments, the average forward walking velocity was set as $\bar{v}^x_{\text{CoM}}=0.1$ to $0.5$~m/s with velocity increments of $0.1$~m/s. We used the dimensionless velocity~\cite{lee2013comparative} to represent the average forward walking velocity, that is, $\bar{v}^{x}_{\text{CoM}}/\sqrt{gh_{\mathrm{CoM}}}$, where $h_{\mathrm{CoM}}$ is the CoM height. Corresponding CoTs of the locomotion were calculated for both on rigid ground and sandy terrain.

\section{Results}
\label{sec:result}

\subsection{Experimental Results}
Fig.~\ref{fig:Locomotion} shows the comparisons of bipedal robot sand-walking locomotion results. We compared the results of the proposed model, the model with the rigid-ground assumptions (i.e., non-slip and non-sinkage), and experimental measurements. {\color{black}Results of representative ten steps are selected and presented over the entire stance phase with the average values and one standard deviation.} Fig.~\ref{fig:q} shows that the proposed model closely followed experiments for both swing and stance legs. Fig.~\ref{fig:FxFyFz} shows the ground-reaction-force estimation and comparisons. For the conventional model, the GRFs were estimated through the inverse dynamics. The Jacobian matrix of the external forces was obtained at the beginning of each stance position and remained unchanged for non-slip and non-sinkage assumptions, leading to overestimated force magnitudes in all directions. Fig.~\ref{fig:xys_sand} shows the intrusion variables between the model prediction and experimental calculation.
Fig.~\ref{fig:rmseVel} further shows the overall one-stance root-mean-square error (RMSE) comparison results under various prescribed moving velocity commands. Fig.~\ref{fig:RMSE_q} compares the error results of the joint kinematics. It shows that the proposed model provided more accurate predictions for hip and knee joint angles, namely, $q_{2,3,5,6}$. Figs.~\ref{fig:RMSE_GRFs} and~\ref{fig:RMSE_xyz} show the RMSE results of GRFs and intrusion, respectively.
%The errors from the proposed model were within $10$ N for the GRF prediction in all directions while the errors of intrusion states were all less than $12$~mm, which was smaller than the foot size ($75\times63\times24$~mm).
The results in Fig.~\ref{fig:rmseVel} confirm that the proposed model offered accurate results for all forward-moving-velocity conditions.

Fig.~\ref{fig:rolling_theta} show the results of the stance foot rolling angle change $\Delta\theta_r$ in the sagittal plane during an entire single stance phase. Fig.~\ref{fig:rolling_Reff} shows the effective rolling radii of the foot on granular media and rigid ground. At the beginning of the stance, the foot started to roll and simultaneously sink with significant change of the orientation and directional angle before $40\%$ of the stance phase. During this period, the effective rolling radius was smooth, implying the rolling contact. In the mid-stance, the orientation angle exhibited a plateau; see Fig.~\ref{fig:rolling_theta}. This implies that the foot was neither pitching rapidly nor significantly reorienting its velocity direction, consistent with a “steady translation slip” phase. After that, the foot continued to roll and sink at the late-stance simultaneously pushing against media backward. However, for rigid-ground walking, the stance foot experienced a large rolling angle $\Delta\theta_r$ nearly without any plateau and the value of effective rolling radius was almost equivalent to the height of CoM.

%Fig.~\ref{fig:Roll_vel} shows the trend of the normalized angle $\bar{\theta}_r$, $\bar{\theta}_r=\frac{\theta_r-\theta^{\min}_r}{\theta^{\max}_r-\theta^{\min}_r}$ as the walking speed increased. An interesting fact is that, the plateau of the rolling angle at low walking speeds tended to disappear as the robot walked faster on the sand terrain, implying that translational slip was replaced by rolling at high-speed walking scenarios. In contrast, the gait of rigid-ground walking remained consistent and showed no velocity dependence.

The CoTs of the robot during the walking on both the rigid surface and sand were calculated. Fig.~\ref{fig:CoT_stance} shows the results of comparisons of CoT calculations at the average forward walking $\bar{v}_{\mathrm{CoM}}^{x}=0.2$~m/s during one stance. Walking on sand requires more energy than walking on the rigid ground. Also, it shows that the proposed dynamics model provided reliable CoT estimation. The average CoT on sand terrain was $2.1$, around $43\%$ more than that on rigid ground (i.e., $1.46$ at the end of the stance phase). The main difference came in the mid- and late-stance, i.e., after $20\%$ of the stance phase. During this period, the value of CoT on sand started to ramp up nearly linearly followed by a significant jump at the late-stance when the foot started to pull off sand. The observed CoT increase on sand was from both terrain-induced energy dissipation and the lack of adaptive strategies given the fixed walking controller under the rigid ground assumptions.

We conducted extensive experiments under various robot forward moving velocities to show the velocity dependency on both the rigid ground and sand. Fig.~\ref{fig:CoT_velocity} shows corresponding CoTs with respect to the dimensionless velocity. We also provided human walking locomotion results computed by the data from~\cite{zhu2025JBE}. It is found that for the bipedal robot, the CoT of sand walking was $1.12\sim1.62$ times more than that of rigid-ground walking. While for human walking locomotion, it took $2.08-2.67$ times more energy than did on the rigid ground. This is consistent with the reported results in~\cite{lejeune1998mechanics}. It is also interesting to find that velocity-dependency of CoT of the bipedal robot shared a similar ‘U’-shaped trend with human locomotion on both terrain conditions.

\subsection{Discussions}

The closed-form granular GRFs contributed marginal computational overhead and the fast real-time implementation with low computational cost is guaranteed given the reference planar robot URDF using Pinocchio~\cite{carpentier2019pinocchio,neuman2019benchmarking}.
This work assumes dry granular media, and variations in particle size and packing fraction are isolated in the GRF model. These variations can be handled via re-identification with representative performance reported in~\cite{chen2025TMECH_Sand}.
With integration of the accurate foot-terrain interaction model to bipedal body dynamics, the proposed framework has potentials for the real-time terrain classification and terramechanics parameter estimation (such as $\alpha_{x,y,z}$ and $\lambda$). This can be done by incorporating the biped morphology and locomotion kinematics with proprioceptive sensors (e.g., encoders and inertial measurement units, i.e., IMUs)using an online identification process.
For cohesive, wet media, additional effects such as adhesion and suction can be incorporated through moisture- or cohesion-dependent terms~\cite{ChenAIM24} while preserving the same augmented walking dynamics.
We considered decoupled, unanchored planar dynamics in sagittal and frontal planes, which was sufficient for straight walking. However, it was not a full 6-DoF floating-base model as the yaw dynamics and 3D coupling of contact forces were not explicitly represented. Extension to a unified 3D augmented dynamics and experimental validations under turning maneuvers are among ongoing future works.

\section{Conclusions}
\label{sec:Cons}

This paper presented a whole-body dynamic model for bipedal walking locomotion on granular media by modeling the intrusion interactions. We relaxed the non-sinkage and non-slip assumptions in the existing models. The model decoupled the sagittal and frontal dynamics and incorporated a new ground-reaction-force model that was specifically developed for foot-granular-terrain interactions. Bipedal walking tests on a sand track were conducted. The results confirmed that the model prediction matched the experiments in both the joint angles and the ground reaction forces. The new whole-body dynamic model can be further used for not only developing model-based optimal-gait design and energy-efficient control, but also computationally effective simulation design for machine-learning-based bipedal robot control and navigation on granular terrains.

\bibliographystyle{IEEEtran}
\bibliography{Chen_Ref}
\end{document}